\documentclass[sigconf]{acmart}

\usepackage{balance}
\usepackage{amsmath}
\usepackage{makecell}
\usepackage{multirow}
\usepackage{pifont}
\usepackage{tabulary}
\usepackage{hhline}
\usepackage{enumitem}
\AtBeginDocument{%
  \providecommand\BibTeX{{%
    \normalfont B\kern-0.5em{\scshape i\kern-0.25em b}\kern-0.8em\TeX}}}

\copyrightyear{2021}
\acmYear{2021}
\setcopyright{rightsretained}
\acmConference[KDD '21]{Proceedings of the 27th ACM SIGKDD
Conference on Knowledge Discovery and Data Mining}{August 14--18,
2021}{Virtual Event, Singapore}
\acmBooktitle{Proceedings of the 27th ACM SIGKDD Conference on
Knowledge Discovery and Data Mining (KDD '21), August 14--18, 2021,
Virtual Event, Singapore}
\acmISBN{978-1-4503-8332-5/21/08}
\acmDOI{10.1145/3447548.3467122}

\begin{document}
\fancyhead{}

\title{Web-Scale Generic Object Detection at Microsoft Bing}

\author{Stephen Xi Chen, Saurajit Mukherjee, Unmesh Phadke \\Tingting Wang, Junwon Park, Ravi Theja Yada}
\affiliation{%
  \institution{Microsoft Corporation}
  \streetaddress{One Microsoft Way}
  \city{Redmond}
  \state{Washington}
  \country{U.S.A.}
  \postcode{98052}
}
\email{{chnxi, saurajim, unphadke, tiwang, junwpar, raviyada}@microsoft.com}
\renewcommand{\shortauthors}{Chen et al.}

\begin{abstract}
In this paper, we present Generic Object Detection (GenOD), one of the largest object detection systems deployed to a web-scale general visual search engine that can detect over $900$ categories for all Microsoft Bing Visual Search queries in near real-time. It acts as a fundamental visual query understanding service that provides object-centric information and shows gains in multiple production scenarios, improving upon domain-specific models. We discuss the challenges of collecting data, training, deploying and updating such a large-scale object detection model with multiple dependencies. We discuss a data collection pipeline that reduces per-bounding box labeling cost by {$81.5\%$} and latency by {$61.2\%$} while improving on annotation quality. We show that GenOD can improve \textit{weighted average precision} by over $20\%$ compared to multiple domain-specific models. We also improve the model update agility by nearly 2 times with the proposed disjoint detector training compared to joint fine-tuning. Finally we demonstrate how GenOD benefits visual search applications by significantly improving object-level search relevance by {$54.9\%$} and user engagement by {$59.9\%$}.
\end{abstract}



\newcommand{\cmark}{\ding{51}}%
\newcommand{\xmark}{\ding{55}}%
\newcommand\tsub[1]{\textsubscript{#1}}
\newcommand{\dset}[1]{$\mathcal{D}_{#1}$}
\newcommand{\model}[1]{$\mathcal{M}_{#1}$}
\newcommand{\tworow}[2]{\textbf{\begin{tabular}[c]{@{}c@{}}#1\\ #2\end{tabular}}}
\newcommand{\threerow}[3]{\textbf{\begin{tabular}[c]{@{}c@{}c@{}}#1\\ #2\\ #3\end{tabular}}}

\maketitle
\section{Introduction}

\begin{figure}[!ht]
\centering
\includegraphics[width=0.4\textwidth]{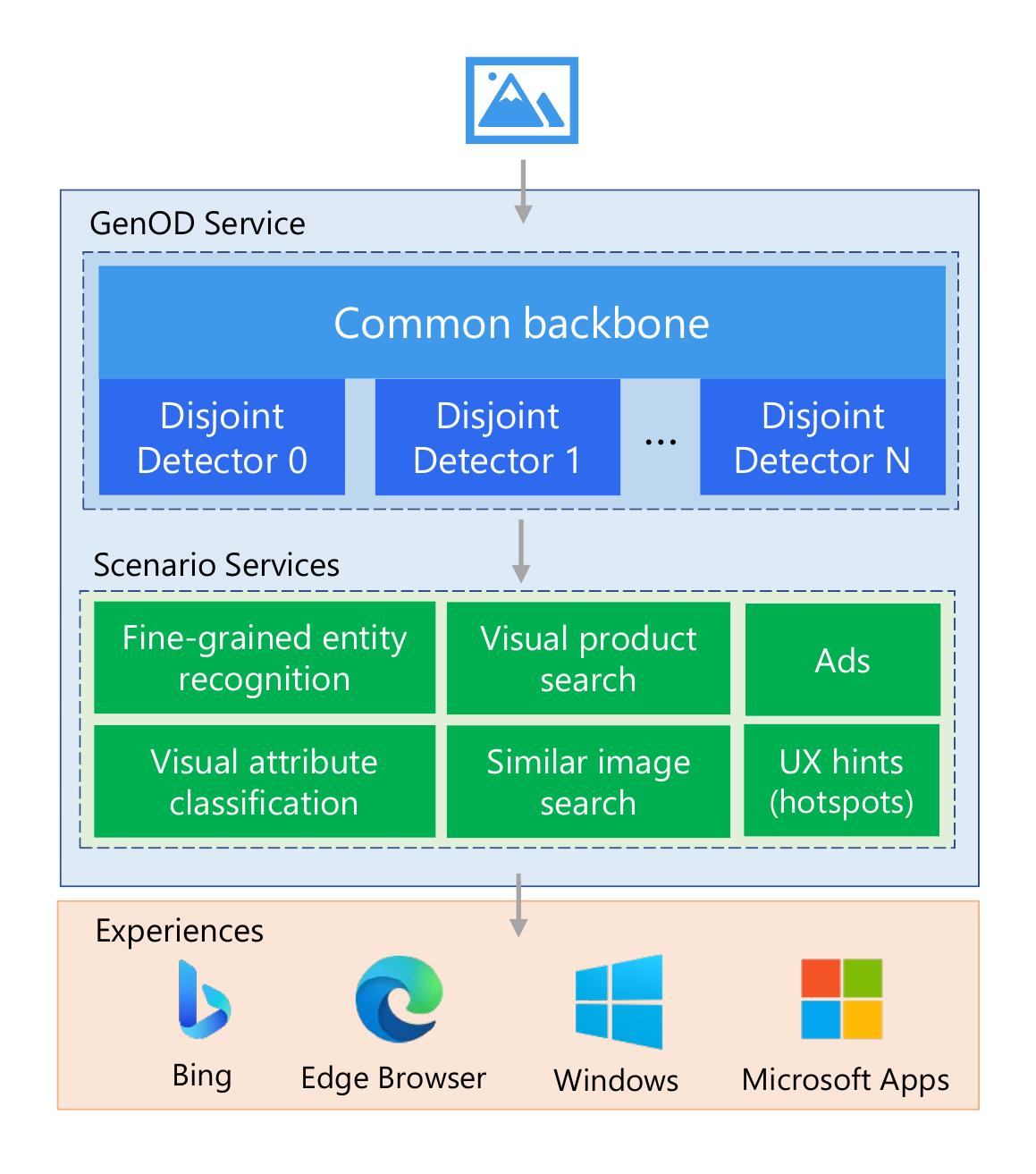}
\caption{Overview of Generic Object Detection (GenOD) in Microsoft Bing Visual Search stack. GenOD is a fundamental image understanding service that provides object level information for a given user query image to trigger multiple scenario services, improve search relevance, and provide user interface hotspots to multiple experience endpoints.}
\label{fig:overview}
\vspace{-3mm}
\end{figure}

Visual search, in which an image constitutes the user query, is an emerging modality of search that allows users to provide a new class of queries beyond text-based search. This search solution requires us to intelligently identify visual concepts, retrieve visually and semantically similar images, search for product information, or get inspiration from other images. Understanding and representing the query image is the critical first step of visual search. Many commercial visual search systems represent query images with image-level embeddings. However, this assumes that the query image is focused on a single object with a clean and simple background which often does not hold true in real world scenarios with mobile captured images.


\begin{figure*}[!ht]
\centering
\includegraphics[width=0.75\textwidth]{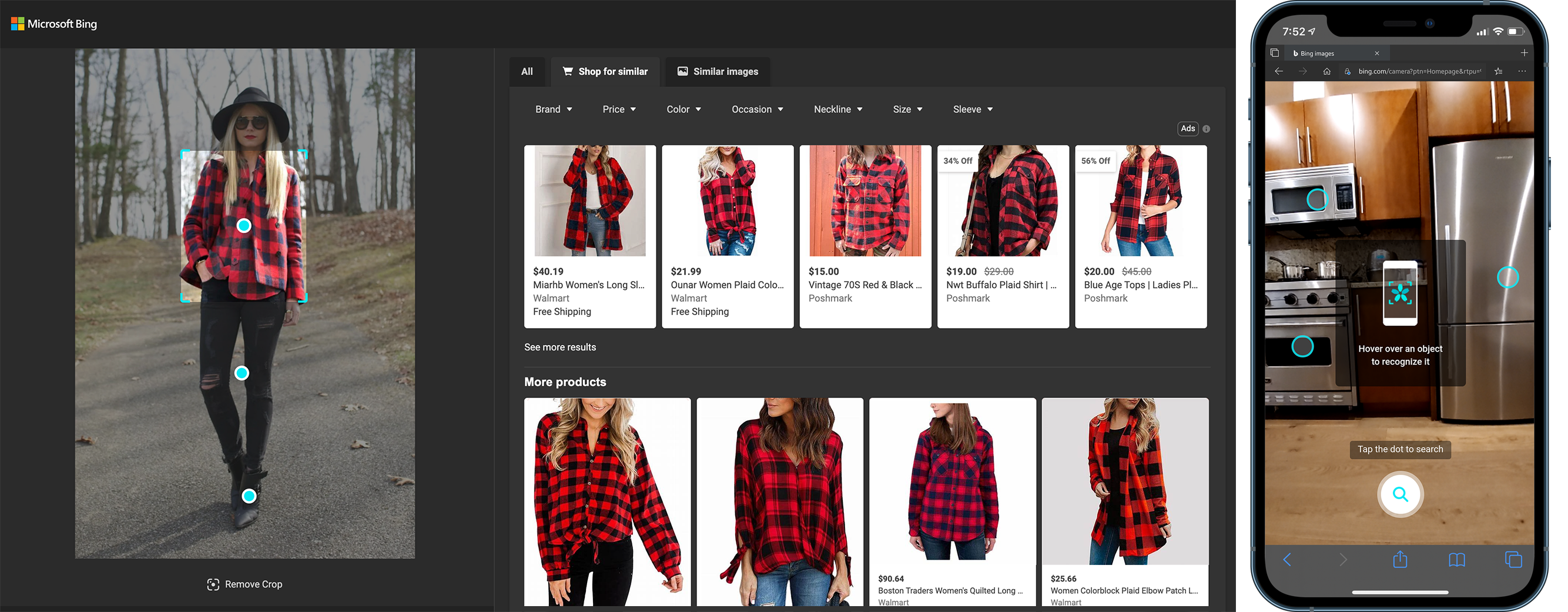}
\caption{Examples of user interfaces with interactive hotspots detected by the Generic Object Detection (GenOD) in the Bing Visual Search experiences. Left: The experience in Bing Image Details Page on desktop, allows users to click on hotspots to search for related products. Right: The Bing Mobile Camera experience detects objects in real-time to allow the user to quickly choose which object is of interest to them.}
\label{fig:applications}
\vspace{-3mm}
\end{figure*}

Object detection has been introduced to several visual search engines~\cite{pinterest_vs_with_od, bingkdd18, groknet, alibaba} to better parse user intent. Given an image, object detection aims to locate and recognize objects from a predefined set of categories. Given their business scenarios, these systems tend to use object detection to display hotspots or remove background noise of objects in scoped segments like shopping. For a web-scale, general-purpose visual search engine like Microsoft Bing, there are numerous search query segments and application scenarios and it is imperative to have a comprehensive and scalable object-based understanding of images at a generic level. 

In this paper, we present how we built Generic Object Detection (GenOD) as a platform service at Microsoft Bing. Figure~\ref{fig:overview} depicts the overview of GenOD in the Bing Visual Search stack. Starting from domain-specific detection models, object detection in Bing has evolved to the generic object level with a taxonomy of over $900$ categories, making it one of the largest deployed object detection models in production. 
With the ability to detect a wide range of objects, GenOD fuels multiple applications including visual product search, image content-based triggering and post-processing, fine-grained entity recognition, fine-grained attribute classification, and real-time object detection in camera search experiences~\footnote{https://bing.com/camera}. Figure~\ref{fig:applications} showcases how users interact with detected object hotspots in Bing Images Details Page and Bing Mobile camera based search experience. The challenges of building such a versatile system can be broken down into three main aspects:

\textbf{Data collection and processing for a large vocabulary} Collecting object detection annotations for hundreds of categories at the scale required for deep models is much slower and more costly than getting image class labels~\cite{su2012crowdsourcing} and can be prohibitively expensive when using expert judges. The task is also fairly complex for crowd platforms, especially because it quickly becomes infeasible for judges to keep track of hundreds of categories. Even if one can leverage recently released large-scale object detection datasets such as OpenImages~\cite{openimagev4}, LVIS~\cite{gupta2019lvis} and VisualGenome~\cite{visualgenome} with hundreds to thousands of categories, determining the best way to combine these resources remains an open issue. Compared to conventional object detection models trained on a single dataset with a small vocabulary~\cite{coco, pascalvoc}, training a unified large-scale detection model by combining several diverse datasets faces new challenges including: (1) \textit{long-tailed category distribution}: this is especially the case in natural images when the number of categories grows 10 times larger. The rare classes often perform poorly with few training samples. (2) \textit{hierarchical labels}: as the taxonomy grows, each object instance naturally has multiple valid labels as part of a hierarchy. For example, an apple can be labeled as "Apple" and "Fruit" because both categories are in the taxonomy. This will introduce missing and noisy label issues because not all object instances can be exhaustively annotated either by humans or oracle models, so it poses a serious barrier in both model training and evaluation. (3) \textit{imbalance between datasets}: Some of the datasets are much larger than others in size with specific distributions, which would be likely dominate model training and cause poorer generalized performance.

\textbf{Agility of model development} 
Continuously iterating machine learning models deployed in online systems remains difficult due to: (1)  \textit{heavy model training}: conventional way of model training is an all-or-nothing change, reducing update agility. (2) \textit{production non-regression requirements}: when a new model is deployed to production, it is important not to regress in performance for any downstream task dependent on the model. However, with the increasing number of categories and dependencies, improving the model for a certain task or subset of categories may lead to a decline in the performance of others, which would block model deployment. Therefore it is imperative to have a novel architecture design to meet such strict requirements. 

\textbf{Latency-accuracy tradeoff} The visual search stack in Microsoft Bing has strict near real-time inference requirements, especially for applications like Bing mobile camera. Since GenOD is required to run for all requests, latency of the model is a key criterion in model training and selection. 

The key contribution of this paper is a detailed description of how we overcome the challenges mentioned above to design and deploy a large-scale generic object detection system in an industry setting that is adaptable to rapidly changing business needs. Specifically, our contributions are as follows: 

\begin{itemize}
\item We discuss the design of a low-cost, high-throughput data collection pipeline that can easily scale to new categories. 
\item We discuss how we handle the imbalance in category and dataset distributions while combining multiple datasets in training a large-scale unified generic object detection model. We evaluate on the various academic and internal benchmarks to demonstrate the efficacy of the model with good speed-accuracy trade-offs and show that a generic large-scale model is able to beat domain-specific models.
\item We propose an architecture design of disjoint detectors on a shared backbone pretrained for general purpose object detection, in order to tackle the challenge of agile model updates in a production context.
\item We describe how we serve GenOD at web scale with low latency, and demonstrate its impact as a fundamental service in Bing Visual Search to improve user engagement and relevance for a wide range of deployed applications in Microsoft through offline and online A/B tests.
\end{itemize}

The rest of the paper is organized as follows: We first review related literature in Section~\ref{sec:related_work} then introduce our data collection pipeline in Section~\ref{sec:data_collection}. Our model design, training and deployment is described in Section~\ref{sec:approach} and we include corresponding experiments in Section~\ref{sec:experiments}. Finally we demonstrate the applications of GenOD in Bing Visual Search in Section~\ref{sec:applications}.

\section{Related Work}
\label{sec:related_work}

\subsection{Object detection in visual search}
Major companies~\cite{glens,amazon,pinterest_vs_with_od,groknet,ebay,bingkdd18,pinterestvisualdiscovery,shopthelookpinterest,alibaba} have been developing visual search systems to satisfy an increasing demand for content-based retrieval. Facebook~\cite{groknet} and Alibaba~\cite{alibaba} perform class-agnostic object localization to remove background noise and retrieve images at object level to improve product search relevance. Pinterest~\cite{pinterest_vs_with_od,pinterestvisualdiscovery,shopthelookpinterest} displays hotspots on objects in a few shopping segments including fashion, home decor and vehicles. ~\cite{bingkdd18} leveraged object detection to improve engagement and relevance in the web-scale responsive visual search system in Microsoft Bing. However, most of these systems target a limited set of shopping domains and only cover a small set of categories. Google Lens~\cite{glens} was one of the first attempts to apply generic object detection for visual search, but a detailed analysis of their system has not been published yet. To the best of our knowledge, this paper is the first work to comprehensively discuss the challenges and solutions for developing a web-scale generic object detection service in a production system. 


\subsection{Large scale generic object detection}
With the advance of deep neural networks (DNN), the research community is moving towards the challenging goal of building a generic object detection system that can detect a broad or even open-ended range of objects like humans~\cite{liu2020deep}. Numerous object detection architectures have emerged during the last decade. \textit{Two-stage detectors}~\cite{ren2015faster, he2017mask} were first proposed to apply DNNs end-to-end to a region proposal network and a detection stage; \textit{one-stage detectors}~\cite{yolov3,lin2017focal,tan2020efficientdet} and \textit{anchor-free} approaches~\cite{law2018cornernet,tian2019fcos} were proposed later with attempts to predict objects without region proposals and anchor boxes, respectively, to further improve speed-accuracy trade-offs~\cite{huang2017speed}. For a more comprehensive survey in the area please refer to ~\cite{liu2020deep}. However, most of these architectures in academic settings seldom consider the agility to add or update categories without regressing others, making them less adaptive in an industry product setting with rapidly changing business needs.

Prevalent works on general purpose object detection are mostly performed on a predefined small set of categories with relatively adequate and balanced training samples (e.g.100$\sim$1000+) for each category~\cite{coco,pascalvoc}. Generic object detection with large vocabulary, in contrast, poses new challenges including long-tail distribution for data collection and model training. Some large-scale datasets~\cite{openimagev4,gupta2019lvis,object365} have been collected to facilitate further research in this scenario, in which challenges and solutions to data collection have been discussed. Recent studies to address the challenges of long-tail distribution include data distribution re-balancing~\cite{mahajan2018exploring, gao2018solution}, class-balanced losses~\cite{lin2017focal}, decoupling representation and classifier learning~\cite{kang2019decoupling, li2020overcoming}. This paper mainly experiments with the data distribution re-balancing approach as a simple but robust baseline, but other directions of research in long-tail object detection could be applied in the future.

\section{Data collection}
\label{sec:data_collection}
In this section, we describe the methodology used to collect data at scale to power the GenOD service. Given the large nature of the vocabulary, it is imperative from a production standpoint to have a robust pipeline that is high quality, cost-efficient, high throughput and agile to taxonomy changes and business needs. Previous iterations \cite{bingkdd18} which relied on ad-hoc collection of data through 3rd party vendors or managed human judges were slow and expensive. We also found that a unified large vocabulary necessitated careful data collection design since it was infeasible for human judges to label images while keeping hundreds of categories in mind. We leveraged crowd platforms to access a large pool of judges for high throughput and cost-efficient labeling. Since crowd platforms are generally not suited for complex annotation tasks, we adapted the orchestration in \cite{gupta2019lvis} for object detection. An overview of the pipeline can be seen in Figure~\ref{fig:data_pipeline}\footnote{Image from https://cutetropolis.com/2016/08/31/links-thats-the-way-they-became-the-brady-bunch by Mike Brailer/ \href{https://creativecommons.org/licenses/by-sa/4.0/}{CC BY-SA 4.0}}. The key stages in the pipeline are described below:
\begin{figure*}[ht]
\centering
\includegraphics[width=0.75\textwidth]{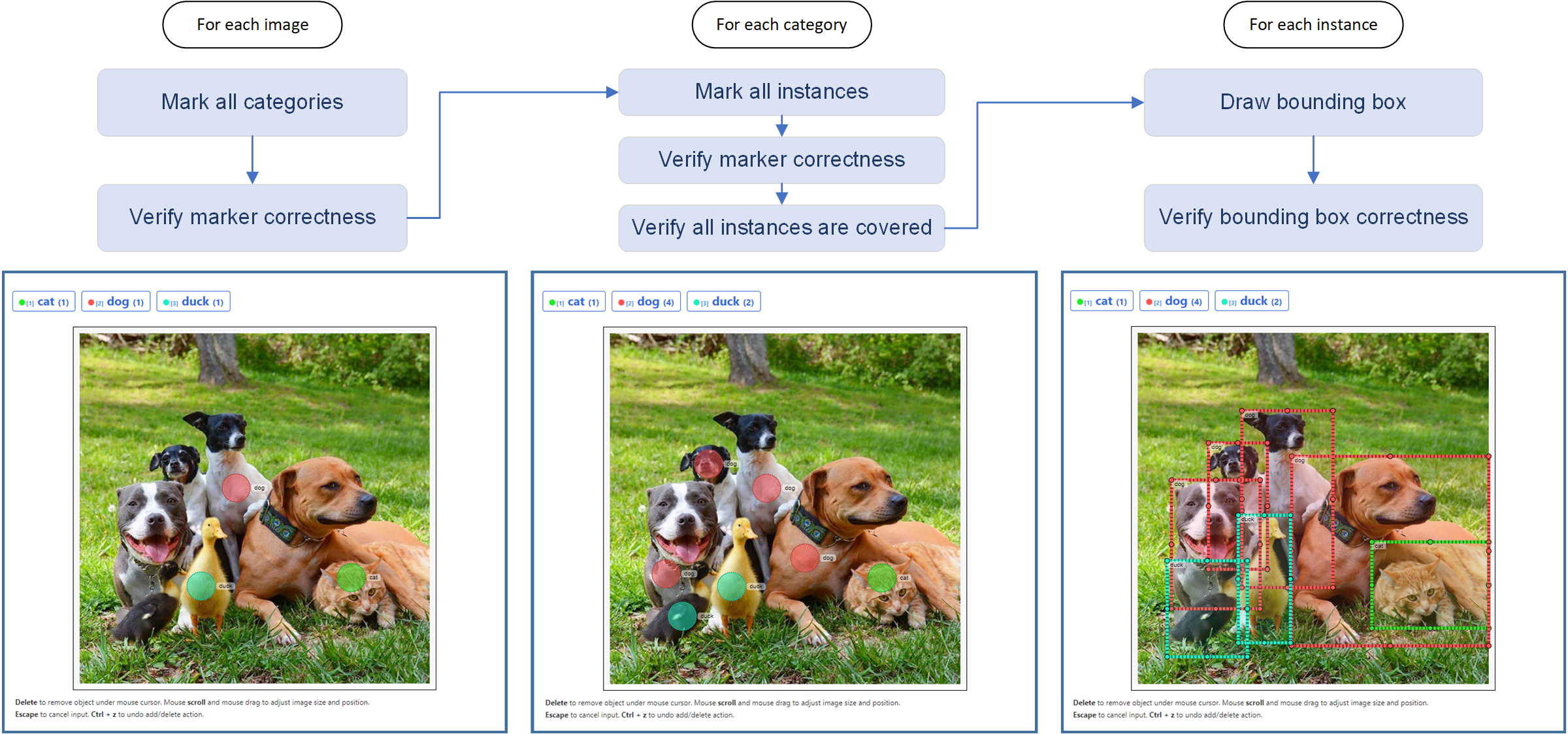}
\caption{The GenOD data collection pipeline is designed as a chain of micro-tasks suited for judging on crowd platforms. It has 3 main stages: category discovery, instance marking and bounding box drawing. Micro-tasks with complex annotations which cannot be easily aggregated are followed by verification micro-tasks with high overlap to ensure quality. The pipeline is orchestrated so crowd judges only have to annotate a single category or marker at a time.}
\label{fig:data_pipeline}
\vspace{-4mm}
\end{figure*}

\subsubsection*{Category Discovery} 
The goal of this stage is to discover all the salient object categories in the image. This is challenging given that there are hundreds of categories and it may be exhausting to label every single object instance in an image. To solve this issue, we ask judges to only discover \textit{a single new category} by placing a marker on an instance or skip if there are no salient object categories to be added (previously marked categories are shown to the judge). This is repeated until 3 consecutive judges skip, at which point we consider all salient object categories have been discovered. We also employ careful user-interface design so the judge can navigate a hierarchy of categories or directly start typing the name of the category to search for the appropriate category. Unlike \cite{gupta2019lvis}, the user interface replaces the cursor with a circle with a size corresponding to the minimum area in an image for salient objects. This ensures judges are not spending time marking insignificant objects that are not important from a user scenario standpoint. We also have a simpler variant of this application where a judge only has to spot a specified category rather than provide the names of new categories. The simpler variant is used when we want to quickly collect data for a single new category for business needs. With these two variants, we are able to quickly discover concepts for our vocabulary while also being agile about adding annotations if the taxonomy expands. After category discovery, we run a marker verification micro-task to ensure that all the marked categories are correct.

\subsubsection*{Instance marking}
The goal of this stage is to mark \textit{all} the instances of the categories discovered in the previous stage. We ask a judge to mark every instance of a given category and follow it up with two quality control micro-tasks: (1) Verify that all instances have been marked (2) Verify all the markers are correct. At the end of this stage, we have markers for all the salient object instances in the given set of images.

\subsubsection*{Bounding box drawing}
The goal of this stage is to draw a bounding box for a given category marker. This is followed up by a bounding box verification micro-task to ensure quality. By decoupling the drawing of the bounding box from the marker, the data collection pipeline is flexible to accommodate future needs such as segmentation.

\subsubsection*{Negative set selection}
The goal of this stage is to collect a set of images for a category such that no instance of that category exist in the images. This stage is not necessary while collecting training data, but is useful for the federated measurement design described in ~\cite{gupta2019lvis}. 

\subsection{Annotation evaluation}
We evaluate the proposed pipeline against the baseline data collection approach which used managed vendor judges. To capture the statistics of camera and web-style images appropriately, we randomly sampled 500 images from each distribution for a total evaluation set of 1k images. When comparing the proposed pipeline's results to the existing baseline annotations, we find that 85.75\% of baseline instances (93.5\% if we exclude objects with smaller dimension < 55 pixels) are correctly localized, and 97\% of the markers are verified as correctly categorized. For a more rigorous comparison that is not biased to the baseline or any particular vocabulary as groundtruth, we sent a subsample of 100 images to expert judges to annotate all salient objects as groundtruth and also verify correctness of bounding boxes provided by each data pipeline. We measure precision for each candidate pipeline's provided bounding boxes and recall against the expert-provided salient bounding boxes. We can see the metrics for quality, cost and latency in Table \ref{tab:datacollcomparison}.



While the throughput at the image level is slightly worse than our baseline approach, this is mainly because our pipeline is more successful at finding more instances to be labeled per image. We ran 2 labeling tasks with 100 and 1000 samples respectively and found that the time taken to get a fully annotated image decreased from 9.3 mins to 4.85 mins. As demonstrated in  ~\cite{lasecki2014using}, task interruption and context switching decreases the efficiency of workers while performing micro-tasks on a crowdsourcing platform. Judges are more likely to work on a task when a lot of data is available to be judged. This suggests that even at the image level, throughput can be increased further by sending larger batches which optimize for the capacity of the crowd platform.
\begin{table*}[h]\small
\begin{tabular}{l|cc|cc|cc|cc}
\toprule
\multirow{2}{*}{}  & \multicolumn{2}{c|}{\textbf{Statistics}} & \multicolumn{2}{c|}{\textbf{Quality}} & \multicolumn{2}{c|}{\textbf{Cost}} & \multicolumn{2}{c}{\textbf{Latency}} \\
                    & \small{\#Categories/img}& \small{\#Bboxes/img}    & \small{Precision}& \small{Recall}   & \small{\$/img}& \small{\$/bbox}      & \small{Time(mins)/img}    & \small{Time(mins)/bbox} \\ 
\hline
\textbf{Baseline}   &2.3 & 3.18     &\textbf{0.959} & 0.602 & 1.89          & 0.65          & \textbf{4.3}  & 0.67           \\
\textbf{GenOD data pipeline}   & \textbf{7.4} & \textbf{14.41}  & 0.933 & \textbf{0.859}   & \textbf{1.63} & \textbf{0.12} & 4.85          & \textbf{0.26}  \\
\bottomrule
\end{tabular}
\caption{Comparison of the GenOD data collection pipeline to the previous method. We find that the new pipeline can discover more object instances, annotate faster, and more cost-effectively per bounding box compared to the baseline.}
\label{tab:datacollcomparison}
\vspace{-5mm}
\end{table*}

\section{Approach}
\label{sec:approach}
In this section, we describe how we developed the GenOD service including data processing to mitigate the serious imbalance in category distribution and dataset sizes, model architecture selection in pursuit of a good speed-accuracy trade-off, and training protocols to achieve a balance between high-performing deep models and model update agility. Our strategy is to first train a single unified base model with a large amount of data to get a generic backbone and have a default detector head for all categories which is easy to maintain and updated less frequently. We improve on this design with the concept of disjoint detectors on the shared backbone, which allows for agile, targeted updates while not disrupting downstream dependencies. Finally we discuss how we deal with the system challenges in scalable serving with low latency.
\subsection{Training Data}
\label{subsec:training_data}
\begin{figure}[ht]
\centering
\includegraphics[width=0.4\textwidth]{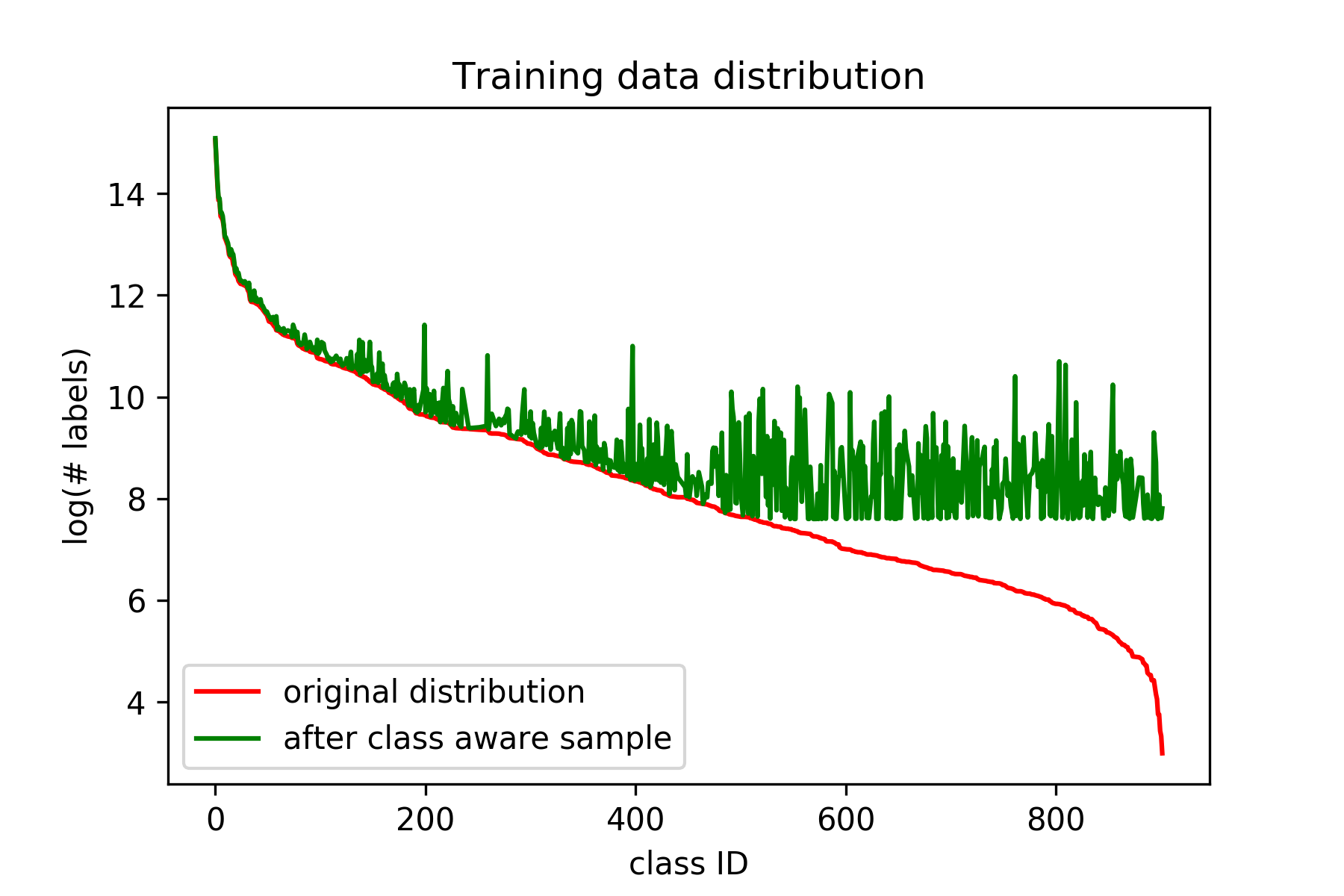}
\caption{Dataset label distribution before and after applying class-aware sampling. We can see the originally skewed distribution becomes more balanced. The class IDs are sorted by the number
of annotated bounding boxes in the original distribution.}
\label{fig:cas2k}
\vspace{-4mm}
\end{figure}

We combine several large-scale object detection datasets such as Bing internal datasets and open source datasets in our training data. These datasets vary from each other greatly in domain (eg. fine-grained commerce versus generic concepts), the number of images, number and hierarchy of categories as well as the coverage and density of objects per image. Therefore combining these heterogeneous datasets for training with a unified taxonomy is a non-trivial task. Another challenge with such a large vocabulary is the long-tailed, extremely imbalanced category distribution, as shown in the red curve in Figure~\ref{fig:cas2k}. Directly training on such imbalanced data would lead to poor performance on the tail categories. There is also an imbalance in the number of images from different datasets, ranging from several millions to a few 100 thousands. Therefore, training would be dominated by the distribution in larger datasets while smaller datasets would be under-represented.

To combine these diverse and imbalanced datasets, we first built a unified GenOD taxonomy with both top-down human design and bottom-up mapping of the categories from all the datasets; organized in a hierarchical manner. To alleviate the poor performance of rare categories with few training samples, we employ a simple yet effective approach of offline class-aware upsampling~\cite{gao2018solution} on the tail classes, during which all the categories will be upsampled to at least $N_{min}$ instances in the training set. In our experiments we use $N_{min}=2000$ as we found it works well empirically. Figure~\ref{fig:cas2k} shows the class-wise distribution in our training set before and after the class-aware sampling. With class-aware sampling, we obtained a total of $3.4$ million training images and $29.3$ million objects. We denote this training set as $\mathcal{D}_{large}$.

To address the imbalance between different datasets, we also downsample the larger datasets offline. This gives us a training set with $1.4$ million images and $10.8$ million objects, which we denote as $\mathcal{D}_{base}$.

\subsection{Base model architectures}
\label{subsec:model_arch}
We have experimented with model architectures including Faster-RCNN~\cite{ren2015faster} and SSD~\cite{liu2016ssd} for shopping-segment object detection models in our previous work~\cite{bingkdd18}. However, with an order of magnitude more categories, in this work we consider the speed-accuracy trade-off as our first priority, and focus on the evaluation of singe-stage detectors which demonstrate better speed-accuracy trade-offs since their inception~\cite{yolov1, liu2016ssd}. We evaluate two variants of single-stage detection models: RetinaNet~\cite{lin2017focal} and FCOS~\cite{tian2019fcos}, state-of-the-art single-stage detectors for anchor-based and anchor-free models respectively at the time of development of GenOD. Both models have achieved good speed-accuracy trade-off at relatively small tasks like COCO. As the number of categories increases, latency for RetinaNet increases dramatically since it has a large number of per-anchor operations and the last few convolution layers to output the class-wise confidence and bounding box regressions become proportionally larger. On the other hand, since FCOS is anchor-free, it reduces the per anchor operations $9$ times compared to RetinaNet. With a few nearly cost-free improvements, FCOS can achieve better results than previous anchor-based detectors with less floating point operations per second (FLOPs) and latency. The experiments in Section~\ref{sec:experiments} provide a comprehensive, quantitative comparison and analysis of these two models.
\subsection{Disjoint detector architecture}
\label{subsec:disjoint}
In a production setting, it is common to have an urgent business need to support a new category or improve a specific category quickly while also not degrading performance on other categories that may have downstream dependencies. With smaller vocabularies, it can be sufficient to retrain the entire model with a new category or more data, but when scaling to a large vocabulary, it becomes very time-consuming to update the entire model and also guarantee no regression in any of the categories. The base model described in the previous section cannot easily accommodate ad-hoc requests or agile updates.

To address this, we designed the GenOD model as a federation of disjoint detector heads that share a fixed common backbone. New detector heads, which include classification and bounding box regression networks, can be trained and added on top of the backbone without disrupting the other detectors. When there is a need to quickly add or update a category, the data collection process described in Section \ref{sec:data_collection} allows us to quickly collect data for that category and then the disjoint principle allows us to update GenOD service with much less data and without disrupting any production dependencies. We explore this through a prototypical experiment in Section~\ref{subsec:exp_disjoint}.

\subsection{Deployment and Serving}
Service latency is an important factor for a core service like GenOD, therefore we deploy the GenOD models to Microsoft Azure GPU clusters. To serve the GenOD models on GPUs, we first convert them to ONNX models and use ONNX Runtime~\cite{ort} as the backend for inference, which provides an $18\%$ speed-up. We built a wrapper of ONNX Runtime on Windows and used the Domain Specific Language (DSL) in~\cite{bingkdd18}, which utilizes a graph execution engine to perform thread-safe inference. To address global scalability issues, we leverage a micro-service system built on Microsoft Service Fabric~\cite{kakivaya2018service} to serve multiple GPU models as micro-services on several global clusters that can scale elastically based on traffic in different regions. Building a cache of detected objects further reduces end-to-end latency. In the end we have built an elastic, scalable GPU serving system for GenOD which can handle hundreds of queries per second across different regions.


\section{Experiments}
\label{sec:experiments}

\subsection{Evaluation metrics and datasets}
\label{subsec:metric_data}
Unless specified, in the following sections, we use \textit{mean average precision} defined in ~\cite{pascalvoc,coco} as our primary metric, where the \textit{average precision} (AP) is calculated as the integral of the area under the precision-recall curve in which detections are considered \textit{true positives} if their \textit{intersection-over-union} (IOU) with the groudtruths are over 50\%. We denote the metric as AP50. AP50 weighs each category equally, however to account for the true distribution of categories seen in production traffic, we also use \textit{weighted mean average precision}@IOU50, denoted as wAP50:
\begin{equation}
\begin{aligned}[rl]
wAP = \frac{\sum_{c\in{\mathcal{C}}} w_c AP_c}{\sum_{c\in{\mathcal{C}}}w_c},\\
\end{aligned}
\end{equation}
where $w_c$ and $AP_c$ are the weight and the AP50 for class $c$ in a validation set of $\mathcal{C}$ classes, respectively. In our setting, we typically set $w_c$ to the number of instances of $c$ in the validation set.
\begin{table*}[!htb]\small
\centering
\begin{tabular}{c|c|cc|cc|cc|cc|cc}
\toprule
\tworow{Architecture/}{Model} & \tworow{Training}{Data} & \multicolumn{2}{c|}{\tworow{Average}{Metrics}} &\multicolumn{2}{c|}{\textbf{COCO 2017}} & \multicolumn{2}{c|}{\textbf{OpenImagesV5}} & \multicolumn{2}{c|}{\tworow{Bing-Fashion}{Val Set}} & \multicolumn{2}{c}{\tworow{Bing-HF}{Val Set}} \\
 &  & AP50 & wAP50 & AP50 & wAP50 & AP50 & wAP50 & AP50 & wAP50 & AP50 & wAP50 \\ 
 \hline
\textbf{RetinaNet} & GenOD $\mathcal{D}_{base}$(1.4M)& 44.88 & 46.30 & 46.44 & 48.99 & 53.63 & 42.66 & 37.27 & 53.81 & 42.16 & 39.73 \\
\textbf{FCOS(GenOD \model{base})} & GenOD $\mathcal{D}_{base}$(1.4M) & \textbf{52.36} & \textbf{55.01}  & {54.95} & {55.69} & 60.67 & 51.34 & {41.01} & {57.84} & {52.82} & {55.17} \\
\textbf{FCOS(GenOD \model{large})} &  GenOD $\mathcal{D}_{large}$(3.4M) & 50.78 & 53.47 & 53.22 & 55.15 & {61.57} & {51.02} & 37.89 & 55.01 & 50.45 & 52.70 \\
\bottomrule
\end{tabular}
\caption{Experiments of GenOD models on the 4 validation sets (OpenImagesV5, COCO 2017, Bing internal fashion and home furniture detection datasets), comparing the RetinaNet and FCOS architectures, and two FCOS variants \model{base} and \model{large}. Overall GenOD \model{base} is selected as the model candidate with the best average AP50 and wAP50 metrics.}
\label{tab:valsets}
\end{table*}

We evaluate candidates on the validation splits of two public datasets (OpenImagesV5\cite{openimagev4}, COCO 2017\cite{coco}) and two of Bing's internal validation sets in fashion and home furnishing, denoted as \textit{Bing-Fashion Val} and \textit{Bing-HF Val}, respectively. We use the average of AP50 and wAP50 metrics over the 4 validation sets as the criteria to select the final model candidate. We then measure final performance on 3 internal test sets: \textit{Bing-Generic Test}, \textit{Bing-Fashion Test} and \textit{Bing-HF Test} which are collected by a weighted sampling of Bing traffic. Note that Bing traffic is much more challenging than the validation data due to a higher proportion of noisy, cluttered scenes in real-world data.
For the COCO dataset, we follow the evaluation protocol in ~\cite{coco} and also report the AP@IOU[0.5:0.95], which is simply denoted as AP. For the OpenImages dataset, we follow the same federated evaluation protocol in the OpenImages challenges~\cite{openimagev4}.


\subsection{Base model training}
\label{sec:base_training}
We implemented both the RetinaNet and FCOS models based on maskrcnn-benchmark~\cite{francisco2019maskrcnn}. Both models are trained with Feature Pyramid Network (FPN)~\cite{fpn} and Focal Loss~\cite{lin2017focal}, using ResNet-101~\cite{he2016deep} as backbone. For FCOS we employ Modulated Deformable Convolution (DCNv2)~\cite{zhu2019deformable} at stage 2$\sim$4 and trained the model with the proposed improvements in~\cite{tian2019fcos} to further boost the accuracy. Both variants are trained for 24 epochs on the dataset in Section~\ref{subsec:training_data} using 8 V100 GPUs,  with a batch size of 64 and learning rate of 0.03. To best optimize for online production latency, we use an input image resolution of 400$\times$667. We use multi-scale training with the shorter side ranging from 300 to 500 while keeping aspect ratios to adapt to different scales of inputs.


\subsubsection{Candidate selection}
Table~\ref{tab:valsets} shows the results of the GenOD models on the four validation sets described in Section~\ref{subsec:metric_data}. During inference, we map the results from the GenOD taxonomy to the corresponding categories in benchmark datasets for a fair comparison. From the table we can see the trained models with FCOS architecture consistently outperform the RetinaNet one. We denote the FCOS models trained with \dset{base} and \dset{large} as \model{base} and \model{large}, respectively. Overall \model{base} achieves the best performance in the aggregated metrics, so we select this model as our candidate for further evaluation.


\subsubsection{Label propagation in the taxonomy hierarchy}
We also experiment with leveraging the hierarchical information in the GenOD taxonomy to propagate the bounding boxes and scores of the fine-grained categories to their ancestors in the taxonomy at inference time. For example, if a "blue jay" is detected, it would also be propagated to generate "bird" and "animal" labels. We select OpenImages as the benchmark because it has a meaningful generic hierarchy. We evaluate label propagation on the two FCOS model candidates trained on GenOD \dset{base} and \dset{large} respectively on the OpenImagesV5 validation sets. Significant improvements have been observed over the original predictions without propagation. Specifically, label propagation improves wAP50 of the \model{base} model from $51.34$ to $61.65$, and improving the performance of the \model{large} model from $51.02$ to $63.16$. Moreover, the AP50 of the \model{large} model is competitive among the best single models in the OpenImages Detection Challenge 2019\footnote{https://storage.googleapis.com/openimages/web/challenge2019.html} that are trained with similar backbones on larger resolutions (800$\times$1333), showing the effectiveness of our model training and post-processing approach.

\begin{table}[ht]
\centering
\begin{tabular}{c|c|cc}
\toprule
\textbf{Model} & \tworow{Label}{Propagation} & AP50 & wAP50 \\
 \hline
\model{base} & \xmark & 60.67 & 51.34\\
\model{large} &\xmark & 61.57 & 51.02  \\
\model{base} & \cmark & 63.27\textbf{(+2.6)} & 61.65\textbf{(+10.31)}\\
\model{large} &\cmark& \textbf{64.44(+2.8)} & \textbf{63.16(+12.14)} \\
\bottomrule
\end{tabular}
\caption{Evaluating label propagation for FCOS model candidates trained on GenOD training data on the OpenImagesV5 validation set.}
\label{tab:label_prop}
\vspace{-4.5mm}
\end{table}

\subsubsection{Comparison with segment models}
Table~\ref{tab:bing5k} shows the test set results of GenOD \model{base} using the wAP50 metric. We compare GenOD \model{base} to two production segment models trained separately for fashion and home furnishing detection using \textit{Bing-Fashion Train} and \textit{Bing-HF Train} datasets respectively. It should be noted that each of those sets is contained within \dset{base}. We find that GenOD \model{base} improves performance on the key product verticals over domain-specific models while significantly reducing the capacity and maintenance cost.


\begin{table}[]\small
\centering
\begin{tabular}{c|c|ccc}
\toprule
 \tworow{Architecture/}{Model} & \tworow{Training}{Data} & \threerow{Bing-}{Generic}{Test} & \threerow{Bing-}{Fashion}{Test} & \threerow{Bing-}{HF}{Test} \\
 \hline
\textbf{RetinaNet} & Bing-Fashion   Train & - & 21.43 & - \\
\textbf{RetinaNet} & Bing-HF   Train & - & - & 27.93 \\
\textbf{RetinaNet} & GenOD   \dset{base} & 28.20 & 20.29 & 32.02 \\
\textbf{GenOD \model{base}} & GenOD   \dset{base} & \textbf{29.63} & \textbf{25.85} & \textbf{35.65} \\
\bottomrule
\end{tabular}
\caption{Evaluation of weighted AP50 on the Bing object detection measurement set.}
\label{tab:bing5k}
\vspace{-4.5mm}
\end{table}

\subsubsection{Latency}
\label{sssec:latency}
In Table~\ref{tab:latency}, we benchmark the latency between the RetinaNet and FCOS variants in the single-GPU and batch-1 setting on the COCO validation sets on V100 GPUs with CUDA 11.0 by averaging the five runs. From the table we can see FCOS is $3.9\times$ faster than RetinaNet. More interestingly, we observed when scaling up from the 80-class COCO taxonomy to the 900-class GenOD taxonomy, RetinaNet becomes nearly $3$ times slower while the latency of FCOS remains stable, which further increases the latency gap between the two models from $1.4\times$ to $3.9\times$. This shows that the last few class-wise convolution layers in an anchor-based model generate significant overhead as the number of categories grows and demonstrating the anchor-free approach is robust in latency against the scaling of vocabulary, making it better suited to large-vocabulary object detection.
\begin{table}[ht]\small
\centering
\begin{tabular}{c|c|c|c}
\toprule
\textbf{Architecture}      & \textbf{GenOD model} & \textbf{COCO model} & \textbf{COCO to GenOD} \\
\#Classes        & $\sim900$          & $80$   &      \textbf{Latency}        \\
\hline
\textbf{RetinaNet}   & $193.4$ms              &    $65.76$ms    & $2.9\times$             \\
\textbf{FCOS} & $49.4$ms             &    $46.9$ms     & $1.1\times$     \\
\hline
{Speedup} & $3.9\times$           &         $1.4\times$  &     \\
\bottomrule
\end{tabular}
\caption{Single GPU batch-1 latency of RetinaNet and FCOS variants of GenOD models on V100 GPU. With the number of categories scaling from 80 classes of COCO to the generic object taxonomy, the speedup of the FCOS architecture grows from $1.4\times$ to $3.9\times$.} 
\label{tab:latency}
\vspace{-4.5mm}
\end{table}
\subsubsection{Experiments on COCO benchmark}
In Table~\ref{tab:coco} we compare the GenOD model to the models trained on COCO with the same architectures.  We can see GenOD consistently outperforms the COCO-trained models especially on small and mid-size objects, even though they target a much broader vocabulary and are not specifically trained for COCO objects.


\begin{table}[ht]\small 
\centering
\begin{tabulary}{0.5\columnwidth}{c|c|cccccc}
\toprule
\textbf{Arch.}& \textbf{\begin{tabular}[c]{@{}c@{}}Training\\ Data\end{tabular}} & \textbf{$AP$} & \textbf{$AP_{50}$} & \textbf{$AP_{75}$} & \textbf{$AP_{s}$} & \textbf{$AP_m$} & \textbf{$AP_l$} \\ \hline
\textbf{FCOS} & COCO & 37.1 & 54.5 & 39.6 & 16.4 & 40.6 & \textbf{56.3} \\
\textbf{FCOS} & GenOD $\mathcal{D}_{base}$ & \textbf{38.3} & \textbf{54.9} & \textbf{41.1} & \textbf{18.0} & \textbf{43.5} & 56.0 \\
\bottomrule
\end{tabulary}
\caption{Comparison of GenOD \model{base} model with the FCOS model trained on COCO on the COCO 2017 validation set.}
\label{tab:coco}
\vspace{-4.5mm}
\end{table}
\subsection{Disjoint detector training}
\label{subsec:exp_disjoint}
As described in Section \ref{subsec:disjoint}, here we compare the conventional joint training approach with our disjoint approach with a prototypical experiment. The baseline is the GenOD model with a jointly trained head for all categories using \dset{base}, i.e, GenOD \model{base}. Given the GenOD \model{base} model, suppose our goal is to improve the \textit{sofa} category in response to user feedback, without performance degradation of other categories within a short development cycle. For the update, we consider an additional set of labeled data: $\mathcal{D}_{update}=\mathcal{D}_{large}-\mathcal{D}_{base}$. Given this additional data, we conduct three experiments and report the results in Table \ref{tab:fed_training}:
\begin{enumerate}[label=(\alph*)]
    \item {\textbf{Joint detector retraining}} \label{itm:joint_baseline}: We train the single-head joint model with all the available data (\dset{large}) using the same training scheme as GenOD \model{base} as described in Section \ref{sec:base_training}. 
    \item {\textbf{Joint detector finetuning}} \label{itm:joint_finetuning}: We randomly sample 50k images from  \dset{update} and finetune the joint detector starting from the GenOD \model{base} model. For this finetuning stage, we use a smaller learning rate of 0.0001 and train on the data for 12 epochs. 
    \item {\textbf{Disjoint detector finetuning}}: We split the GenOD \model{base} head to create a disjoint detector head for just the \textit{sofa} category. We finetune this model on the same dataset (50k randomly sampled images from \dset{update}) as described in  \ref{itm:joint_finetuning} above. We use a learning rate of 0.00003 and train the disjoint detector head of the model for 12 epochs.
\end{enumerate}

As seen from the experimental results in Table \ref{tab:fed_training}, disjoint detector finetuning on just a small amount of data is far more agile and allows us to train $\sim$\textbf{300x} faster than joint retraining in \ref{itm:joint_baseline} and $\sim$\textbf{2x} faster than  joint finetuning in \ref{itm:joint_finetuning}, while also improving on the category AP. This is achieved without disrupting the existing model for any of the other categories which allows for stable updates in the production stack compared to the conventional model retraining process. 




\begin{table*}[htb]\small
\centering
\begin{tabular}{l|cccccc}
\toprule
\textbf{Description} & \tworow{Training}{Methods} & \tworow{Initialization}{Model} & \tworow{Training}{Data}  & \tworow{\#Training}{Images} & \tworow{Training}{time} & \textbf{AP\tsub{sofa}} \\ 
\hline
Baseline: GenOD \model{base} & Joint & ImageNet & $\mathcal{D}_{base}$ & 1.4M & 8 days & 0.6453 \\ 
\hhline{=|======} 
(a) Joint detector retraining   & Joint & ImageNet & $\mathcal{D}_{large}=\mathcal{D}_{base}$ + $\mathcal{D}_{update}$ & 3.4M & 19 days & 0.6254 \\
(b) Joint detector finetuning & Joint & GenOD \model{base} & 50k random sample from \dset{update} & 50K & 188 mins & 0.6443 \\
(c) Disjoint detector finetuning & Disjoint & GenOD \model{base} & 50k random sample from \dset{update} & 50K & \textbf{100 mins} & \textbf{0.6499} \\
\bottomrule
\end{tabular}
\caption{Evaluation of model update agility for the \textit{sofa} category. Disjoint training of the targeted category is much faster, while also increasing its AP and keeping other categories stable.}
\label{tab:fed_training}
\vspace{-4.5mm}
\end{table*}

\section{Applications}
\label{sec:applications}
\subsection{Object Detection for Visual Product Search}
One of the primary applications for GenOD is to help users better formulate visual search queries and improve the relevance of search results. Figure~\ref{fig:applications} showcases the hotspot interactions in the Bing Image Details Page. GenOD assists the user in formulating a query. Instead of the user having to manually crop to region of interest in the image, detected hotspots are shown over the image for users to start their search with just a tap/click. GenOD's detected categories can be passed to downstream tasks like similar image search ranker to filter out semantically irrelevant results and improve relevance. 
We trained and deployed a lightweight category classifier to the index images. The detected category of the query will be sent to match with the categories of the index images, and filter out those images that do not match.

We conducted comprehensive offline and online experiments on the efficacy of GenOD improving visual search experience. We measure the \textit{defect rate} of the top-5 visual search results from the hotspot clicks on the fashion segment, where defect rate is defined as the average error rate in the categories of the retrieved images. In table~\ref{tab:defect_rate} we can see that after applying the GenOD categories for filtering, the hotspot click-through defect rate has dropped by $54.9\%$, substantially improving the relevance and user experience. 
\begin{table}[!htb] 
\begin{tabular}{cc}
\toprule
\textbf{Filtering approach} & \textbf{Defect rate@5} \\ \hline
{W/o object category} & 38.27\% \\ 
{Object category filtering} & 17.26\% (\textbf{-54.9\%}) \\ 
\bottomrule
\end{tabular}
\caption{Defect rates of top-5 visual search results from object detection hotspot clicks (lower is better). With the ranking results filtered semantically by object detection categories, the product search defect rate decreases significantly by 54.9\%. }
\label{tab:defect_rate}
\vspace{-4mm}
\end{table}
\begin{table}[!htb]
\begin{tabular}{cc}
\toprule
\textbf{Engagement metrics} & \textbf{Gains} \\ \hline
\%Users entry to Visual Search & +14.03\% \\
Overall entry to Visual Search & +27.89\% \\
Hotspot clicks per user & +59.90\% \\ 
\bottomrule
\end{tabular}
\caption{Aggregated online user engagement metrics after deploying GenOD to Bing Visual Search, which shows significant gains over baseline.}
\label{tab:online_engagement}
\vspace{-4mm}
\end{table}

We also set up a series of online A/B tests to measure the user engagement before and after deploying GenOD to all Bing Visual Search requests, as shown in Table~\ref{tab:online_engagement}. After aggregating the online user engagement metrics including the percentage of user entries to visual search, overall entries to visual search and hotspot clicks per unique users, GenOD shows significant gains in bringing in more engaging users in visual search; demonstrating the advantage of expanding object detection to generic objects on all images.

\subsection{Object-Level Triggering for Fine-grained Visual Recognition}
Bing Visual Search runs multiple fine-grained recognition models including animals, plants, landmarks and more. Since these models usually have high latency, it is necessary to perform lightweight triggering before running them. Previous image-level visual triggering models such as the one described in ~\cite{bingkdd18} often fail to trigger on small objects or when multiple different objects are in the scene. For a fair comparison with the previous approach, we compare triggering performance at the image level by aggregating the outputs of the GenOD model. In Table~\ref{tab:food_trig} we evaluate the image-level triggering precision and recall on the food recognition measurement set. For triggering, we prefer models with higher recall performance. Compared to the image-level triggering model, GenOD improves the triggering recall by detecting and recognizing smaller objects.
\begin{table}[!htb] 
\centering
\begin{tabular}{c|cc}
\toprule
\textbf{Triggering approach} & \textbf{Precision} & \textbf{Recall} \\ \hline
Image-level & 99.8 & 81.3 \\
Object-level & 98.5 & 88.1 (\textbf{+8.36\%}) \\
\bottomrule
\end{tabular}
\caption{Comparison of triggering precision and recall between image-level and object-level triggering on the food recognition measurement set.}
\label{tab:food_trig}
\vspace{-4mm}
\end{table}


\subsection{Bing Mobile Camera}
Bing mobile camera is a real-time mobile web experience allowing users to search without having to manually capture a picture. The mobile interface is shown in Figure~\ref{fig:applications} (right). When a user opens the camera experience and phone is stable, a frame will be captured and sent to the GenOD service to perform real-time object detection. The detected objects will be sent back to the phone to render as hotspots. An in-app object tracking model keeps track of the objects when they are in view, so hotspots can stay on the objects without the need to make additional GenOD requests. Clicking on the detected hotspots provides relevant results to the selected object. GenOD enhances the user experience and simplifies the formulation of visual search queries. Online A/B tests show a reduction of 31.8\% in responses with no visual search results compared to our control in which we depend on users to formulate a visual query.

\section{Conclusions}
We presented GenOD, a web-scale generic object detection service that is fundamental in image understanding for Bing Visual Search and fueling multiple downstream applications. We described an efficient data collection workflow for training the models used in the system. We demonstrated with experiments that the GenOD model has achieved competitive object detection results across production measurement sets and academic benchmarks, with good speed-accuracy tradeoff, and can be updated with agility while being stable to dependencies. Specifically, we have shown that by moving to a large-scale single unified generic detector, GenOD can achieve better results than multiple domain-specific models in each vertical, reducing the cost of maintaining several segment models. Finally we also showed how GenOD benefits visual search applications by significantly improving user engagement and search relevance.



{\small
\begin{acks}
We would like to thank Arun Sacheti, Meenaz Merchant, Surendra Ulabala, Mikhail Panfilov, Andre Alves, Kiril Moskaev, Vishal Thakkar, Avinash Vemuluru, Souvick Sarkar, Li Zhang, Anil Akurathi, Vladimir Vakhrin, Houdong Hu, Rui Xia, Xiaotian Han, Dongfei Yu, Ye Wu, Vincent Chen, Kelly Huang, Nik Srivastava, Yokesh Kumar, Mark Bolin, Mahdi Hajiaghayi, Pengchuan Zhang, Xiyang Dai, Lu Yuan, Lei Zhang and Jianfeng Gao for their support, collaboration and many interesting discussions. 
\end{acks}
}

\bibliographystyle{ACM-Reference-Format}
\balance
\bibliography{bibliography}

\end{document}